# Classification of Geographical Land Structure Using Convolution Neural Network and Transfer Learning

[1]**Mustafa Majeed Abd Zaid,** [1]**Ahmed Abed Mohammed and** [2]**Putra Sumari**

[1]*College of Technical Engineering, Islamic University, Najaf, Iraq*
[2]*School of Computer Science, University Sains Malaysia, Penang, Malaysia*

**Abstract:** Satellite imagery has dramatically revolutionized the field of geography by giving academics, scientists, and policymakers unprecedented global access to spatial data. Manual methods typically require significant time and effort to detect the generic land structure in satellite images. This study can produce a set of applications such as urban planning and development, environmental monitoring, disaster management, etc. Therefore, the research presents a methodology to minimize human labor, reducing the expenses and duration needed to identify the land structure. This article developed a deep learning-based approach to automate the process of classifying geographical land structures. We used a satellite image dataset acquired from MLRSNet. The study compared the performance of three architectures, namely CNN, ResNet-50, and Inception-v3. We used three optimizers with any model: Adam, SGD, and RMSProp. We conduct the training process for a fixed number of epochs, specifically 100 epochs, with a batch size of 64. The ResNet-50 achieved an accuracy of 76.5% with the ADAM optimizer, the Inception-v3 with RMSProp achieved an accuracy of 93.8%, and the proposed approach, CNN with RMSProp optimizer, achieved the highest level of performance and an accuracy of 94.8%. Moreover, a thorough examination of the CNN model demonstrated its exceptional accuracy, recall, and F1 scores for all categories, confirming its resilience and dependability in precisely detecting various terrain formations. The results highlight the potential of deep learning models in scene understanding, as well as their significance in efficiently identifying and categorizing land structures from satellite imagery.

## Introduction

Satellite imaging has significantly transformed the discipline of geography by providing academics, scientists, and policymakers unparalleled access to spatial data on a worldwide level (Burke *et al*., 2021). Geographic Information Systems (GIS) utilize satellite images to generate maps, analyze geographical trends, and simulate environmental processes (Chuvieco, 2020). Researchers leverage satellite data to assess changes in land cover and land use, monitor deforestation, track urban expansion, and evaluate the impacts of natural disasters such as floods, wildfires, and earthquakes. In urban planning and changes and guiding decision-making processes (Zhu *et al*., 2019). Urban planners analyze population density, detect land use trends, strategize transit networks, and assess the environmental consequences of urbanization (Koroso *et al*., 2021). High-resolution satellite images facilitate an in-depth examination of urban morphology and agricultural practices. Farmers and agronomists utilize satellite data to evaluate crop conditions, monitor plant development, identify pest infestations, and optimize irrigation and fertilization techniques. Remote sensing technologies provide critical information on soil moisture levels, crop stress conditions, and yield forecasts, enabling farmers to enhance productivity, reduce input costs, and mitigate environmental impacts. Despite the advancements in satellite technology, traditional methods of categorizing geographical land formations often rely on subjective analysis by human experts (Tsatsaris *et al*., 2021).

Conventional methods usually need humans to analyze satellite photos manually, which may be a time-consuming and labor-intensive process. Analysts frequently employ techniques such as visual interpretation, pixel-based categorization, and object-based picture analysis. Nevertheless, these methods possess notable constraints. Firstly, their subjectivity and reliance on manual interpretation make them susceptible to human fallibility and prejudice, resulting in inconsistencies in classification outcomes. Secondly, the laborious process of manually analyzing extensive datasets can be time-consuming, thereby impeding prompt decision-making. Lastly, scalability issues may arise, hindering the application of these methods to larger datasets. With the growing amount of satellite imagery, traditional methods are unable to keep up, making it difficult to analyze large areas efficiently. Moreover, these methods may not accurately capture complex land cover types or subtle variations in land structures, leading to lower classification accuracy.

More effective and trustworthy approaches to land structure categorization are required in light of these difficulties. There is a chance to automate and improve the accuracy of land structure categorization thanks to recent advances in computer vision and deep learning. A subset of machine learning known as "deep learning" trains artificial neural networks to process data by identifying patterns and then making predictions (Wang *et al.*, 2023). Deep learning models are superior to traditional ML techniques in that they can learn to build hierarchical data representations from scratch, ignoring the necessity for human-created features. An example of a CNN is ResNet-50, while another is Inception-v3, which offers an effective means for categorizing terrain formations using satellite imagery. By autonomously extracting features from unprocessed satellite images, these models can effectively distinguish between various types of land cover, significantly aiding environmental monitoring, land management, and urban planning.

The objective of this article is to develop a sophisticated deep-learning model capable of accurately categorizing various geographical land formations in Malaysia through the analysis of satellite imagery. This study aims to evaluate the performance of several advanced architectures, including CNN, ResNet-50, and Inception-v3, to identify the most efficient approach for land structure classification and reduce the time and labor required for land structure identification, making it more efficient and scalable. Ultimately, the goal is to provide a reliable and efficient solution for automated scene understanding, with significant applications in land use planning, environmental conservation, and disaster management.

*Related Work*

In order to improve traffic management and road safety, this study introduces a classification approach that makes use of Convolutional Neural Networks (CNNs) and transfer learning. The system makes use of a dataset of 7616 pictures and uses a CNN architecture with 24 convolution layers and 8 fully linked layers. With a 98.9% accuracy rate, InceptionV3 was the top-performing model throughout the validation phase. In both the validation and test stages, the CNN model achieved an accuracy of 95.1% (Abed Mohammed and Sumari, 2024).

The authors suggest a method for identifying High-Spatial-Resolution Remote Sensing (HSRRS) scenes dubbed TL-DeCNN, which makes use of deep convolutional neural networks. This method effectively prevents overfitting during training and makes use of transfer learning and fine-tuning on a small number of HSRRS scene samples. The experimental findings indicate that the TL-DeCNN approach surpasses the directly trained VGG19, ResNet50, and InceptionV3 models when applied to few-shot data. This technique achieves greater results without experiencing overfitting. Additional investigation might investigate the fundamental principles that contribute to InceptionV3's superior performance in mitigating overfitting and gradient disappearance (Li *et al.*, 2020).

An article published in 2021 focuses on the issue of remote sensing scene categorization, which is crucial for analyzing land cover. The design presents a new encoding method called multi-granularity neural network encoding. It utilizes the models InceptionV3, InceptionResNetV2, VGG16, and DenseNet201. The strategy improves performance and reduces processing costs by using pre-trained Convolutional Neural Networks (CNNs) and ensemble learning approaches. The experimental findings obtained from both public and custom datasets clearly indicate that fine-tuning deep convolutional neural networks for remote sensing applications yields much higher accuracy compared to standard approaches, hence demonstrating their superior efficacy. The suggested fine-tuned complete pre-trained model achieves superior results on the test set and achieved a high level of accuracy of 97.84%, compared to the features retrieved using InceptionResNetV2 (Bosco *et al.*, 2021).

Given the large number of remote sensing pictures accessible for Earth observation and land monitoring, there is an urgent need for sophisticated analytic methods to achieve precise Land Use (LU) categorization. In order to address the challenges associated with LU classification, the purpose of this research is to provide a novel architecture known as AMUSE-CNN, which stands for adaptive multiscale superpixel embedding convolutional neural network. AMUSE-CNN uses a Multiscale Convolutional Neural Network (MS-CNN) to categorize pictures using superpixels of different sizes after using a superpixel representation for object-based analysis. Two modules are in charge of executing the architecture for complete LU classification and an adaptive approach improves the ability to classify. By comparing it to the state-of-the-art methods, the method's superior performance is shown using remote sensing data from Nigeria's Kano and Lagos (Zhang *et al.*, 2022).

Models trained using ensemble learning approaches are more robust and accurate than those trained using individual models. Because of its effectiveness in combining cyclic learning rate schedules to capture the ideal model in each cycle, the snapshot ensemble Convolutional Neural Network (CNN) finds significant usage in many domains. In this research, the dropCyclic rate of learning scheduling is presented as a tool for investigating various local minima. The learning rate is reduced incrementally in each epoch using this step decay approach. We compare three foundation CNN designs (MobileNetV2, VGG16, and VGG19) and three learning rate schedules (dropCyclic, max-min cyclic cosine, and cyclic cosine annealing) on all three aerial image data sets (UCM, AID, and EcoCropsAID) to determine the effectiveness of the snapshot ensemble Convolutional Neural Networks. Szegedy *et al.* introduced the inception model, a deep Convolutional Neural Network (CNN) architecture built for GoogleNet, at the 2014 large-scale image net visual identification challenge. The dropCyclic method is definitely better, according to the findings. It proves its effectiveness in improving the performance of snapshot ensemble CNN by showing higher classification accuracy than other methods Noppitak and Surinta (2022).

In order to determine if it is possible to integrate three separate industries in rural regions, this research used deep learning and AI clustering approaches. When used in conjunction with the k-means algorithm, the ResNet-50 model outperforms competing systems by a margin of 3.1%, achieving a success rate of 88.3% in land-use classification and identification. The model provides policymakers with a helpful tool for fostering the integration of rural industries and designing successful development plans, with an average IoU of 67.29% (Huang *et al.*, 2023a).

This study presents a new categorization method that utilizes a collaborative decision-making process, including many structures and pre-trained Convolutional Neural Networks (CNNs). Three Convolutional Neural Networks (CNNs), namely AlexNet, Inception-v3, and ResNet18, are used individually for land use categorization. The final classification outcomes are established by using a collaborative decision-making approach. The technique entails the creation of novel, fully connected Softmax classification layers, the training of the created CNNs, and the amalgamation of their prediction outcomes. The suggested technique outperforms state-of-the-art methodologies, as shown by the evaluation conducted on UC Merced, AID, NWPU-45, and OPTIMAL-31 datasets (Xu *et al.*, 2020).

Using hyperspecific satellite images from the GaoFen-5 (GF-5) spacecraft, this research evaluates and analyses the comprehensive UFZ categorization in Wuhan, China. Using both traditional classification methods and more recent deep learning techniques, we compare the performance of hyperspectral (GF-5) and multispectral (Landsat 8) data and find that hyperspectral data is far better. The proposed system, SSUN-CRF, integrates spectral and spatial data using unified networks built using deep learning methods. Additionally, it includes a conditional random field that is completely linked. The Underground Freezing Zone (UFZ) may be precisely mapped using this approach. The approach produced outstanding outcomes, with a total accuracy of 93.86% and a Kappa value of 92.08%. (Yuan *et al.*, 2022).

This study presents a comprehensive land-cover map output for Wuhan and its surrounding areas that makes use of the Low-to-High Network (L2HNet). To expedite the mapping procedure, we enhance the efficiency of L2HNet through the elimination of specific components. Google Maps offers remote sensing images at a high resolution, while ESA LandCover (2021) provides categorization at a lower resolution. When the Mean of the Intersect over Union (MIoU) exceeds 75.21%, the product maps with a resolution of precisely 1 m. Additionally, it frequently achieves accuracy rates exceeding 85.00% for Frequency-Weighted Intersecting over Union (FWIoU), OA, and Kappa (Huang *et al.*, 2023b).

This article introduces a methodology for identifying unfamiliar patterns of airport distribution across several study domains by using deep learning and spatial analysis. The approach obtains a high recall rate of 96.4% and an airport integrity rate of 97.2% in a sample area of 21,9040.5 km$^2$ by training a scene classification model on Google image data and including geographical data like road networks and water systems (Li *et al.*, 2021).

This article introduces a novel approach to ensemble learning termed semi-MCNN, which utilizes multiple-CNN models in a semi-supervised manner. Semi-MCNN employs a semisupervised learning approach to use abundant unlabeled data effectively. It does this by autonomously picking samples and creating a dataset. To minimize the spread of mistakes, the technique addresses them by first training on carefully chosen unlabeled data and then refining the training on labeled data. The incorporation of semisupervised learning into a multi-CNN ensemble architecture improves the capacity to generalize and increases the accuracy of classification (Fan *et al.*, 2020).

Researchers proposed a new Convolutional Neural Network (CNN) model that incorporates a triple feature fusion pattern to improve Land Cover Zone (LCZ) detection. This method links many layers of cascaded information to make category judgments, reducing the loss of valuable feature information. The results obtained from the So2sat LCZ 42 dataset show substantial improvements in both overall accuracy and kappa coefficient when compared to sophisticated LCZ classification methods. The suggested model demonstrates an accuracy of 0.70 and a kappa value of 0.68, indicating improvements of about 4.47-6.25%, respectively (Ji *et al.*, 2023).

This study aims to fill the knowledge gap on land use in Morocco by using six machine-learning methods to analyze Landsat 8 satellite data. Using the Google Earth Engine, the machine learning methods employed in this study, the following machine learning algorithms are included: (SVM), (RF), (CART), (MD), (DT), and (GTB). This research evaluates the performance of many techniques. The MD algorithm has the highest accuracy of 0.93, while SVM has the lowest accuracy of 0.74. The inclusion of indices such as NDVI, NDBI, BSI, and MNDWI enhances the total accuracy, with MD showing an increase of about 93%. MD stands out as the most effective classifier, especially when dealing with difficult terrain (Ouchra *et al.*, 2023).

Using satellite images, this essay discusses deforestation as a Multilabel Classification (MLC) problem. The ForestViT model employs self-attention to identify deforestation, eliminating the requirement for convolution operations often used in traditional deep learning models. ForestViT's performance in Multilabel Classification (MLC) has been shown to be promising via experimental assessment of accessible satellite imagery datasets (Kaselimi *et al.*, 2023).

Via high-resolution multispectral WorldView-3 satellite photos, this research presents a technique that accurately delineates individual tree tops via marker-controlled watershed segmentation. The procedure of gradient binarisation precisely identifies the boundaries of tree crowns, using a supervised searching process to establish the appropriate threshold for binarisation. The spatial local maxima, which serve as markers in watershed segmentation, are further improved to remove spurious treetops (Tong *et al.*, 2021).

Classification of land use and land cover using deep learning is a novel approach that is introduced in this research. Locating different kinds of land cover in Remote Sensing Images (RSIs) using the River Formation Dynamics Algorithm (LULCC-RFDADL). Using RFDA to fine-tune hyperparameters, the LULCC-RFDADL model extracts features using the very small EfficientNet. Classification is performed using the Multi-Scale Convolutional Autoencoder (MSCAE) model and the Search Optimization Method (SOA) is employed to optimize parameter selection. According to benchmark dataset analysis, LULCC-RFDADL performs better than competing approaches on several metrics (Aljebreen *et al.*, 2024).

## Materials and Methods

The methodology consists of a set of phases: Dataset collection, preprocessing, classification model, optimization models, performance metric, and selecting the best model as shown in Fig. (1).

### Phase 1: Dataset Description

The satellite picture collection was acquired from MLRSNet, a high-resolution remote sensing dataset that has been particularly created for the purpose of semantic scene comprehension, with the ability to identify and classify various objects or features within the image. Qi *et al.* (2020). MLRSNet offers diverse perspectives of the world captured by satellites, providing high-resolution imagery. The dataset comprises a total of 46 image categories, each with an image size of 256×256 pixels. However, for this specific experiment, we focused on four distinct categories: Farmland, terrace, meadow, and desert. These categories were selected for their relevance to the experiment's objectives. As previously mentioned, we utilized a dataset with four distinct labels, each containing approximately 2600 images, the results of which can be visualized in Fig. (2).

### Phase 2: Preprocessing

We will explain all the preprocessing that will be done on our dataset.

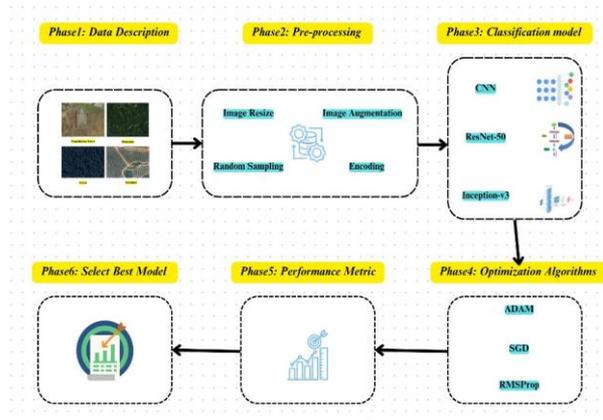

**Fig. 1:** Methodology of the study

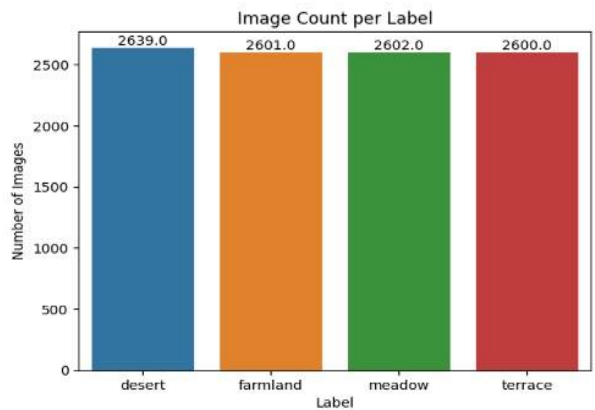

**Fig. 2:** Image count per label in the dataset

*Image Resize*

It is essential preprocessing to change the size of the image and its dimensions while maintaining its aspect ratio or stretching it to fit a new size. To ensure uniformity and compatibility, all photos in our dataset were uniformly downsized to a resolution of 224×224 pixels. This resizing procedure guarantees that all images possess identical dimensions, hence allowing consistent processing and analysis. Standardizing the picture size enables us to efficiently use machine learning algorithms and methodologies to extract significant features and patterns from the information. Figure (3) shows a few examples from the datasets based on desert, farmland, meadow, and terrace and the images' size is 224×224 pixels.

*Image Augmentation*

To expand the dataset further, we employed image augmentation techniques. Image augmentation is an often-used approach to enhance the general-isolation Model performance by introducing variations to the data (Shorten and Khoshgoftaar, 2019). Figure (4) illustrates various image augmentation and transformations, such as rotation, flipping, shearing, and more, which are done by one of the" terrace" images. By applying image augmentation, we increased the number of images for each label to ensure a balanced dataset. In this case, all labels were augmented to have a consistent count of 3500 images. Hence, there will be a total of 14,000 images to form the dataset. This augmented dataset enables the method to obtain from a broader scope of variations and improves its ability to generalize and make accurate predictions.

*Encoding*

It is a technique used to transform values from one representation to another. In the context of categorical variables, label encoding, also known as ordinal encoding, assigns a unique integer value to each category within the dataset (Potdar *et al.*, 2017). This process allows for converting categorical labels into a numerical format that machine learning algorithms can process. In this study, label encoding has been applied to the labels within the dataset, enabling the model to interpret and learn from the encoded representations during the training and prediction stages. Decoding will be applied during visualization to enhance result interpretation, restoring the encoded labels to their original categorical form for easier understanding of the outcomes.

*Random Sampling*

It is a widely used technique for creating a representative subset of a dataset (Gemulla, 2008). It is beneficial for dividing the data into training and testing sets. Another related method is stratified sampling, a probability sampling approach that considers the distinct groups or categories within the dataset; stratified sampling ensures that the number of sampled items from each group is proportional and balanced. In this article, random sampling was employed to split the dataset of 14,000 images into training, testing, and validation sets. The training set consists of 8,400 images (60%), the testing set contains 4,200 images (30%) and the validation set comprises 1,400 images (10%). Random sampling ensures that the selected samples represent the overall dataset and helps maintain the integrity of the data during model evaluation and testing.

*Phase 3: Classification Model*

We will explain all the models that will be used in this study.

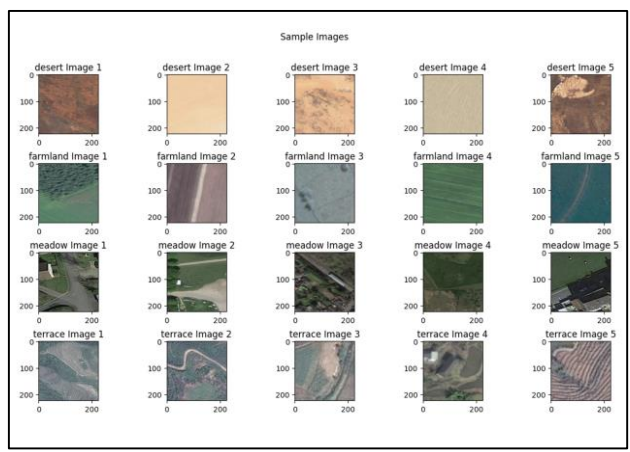

**Fig. 3:** Samples of image

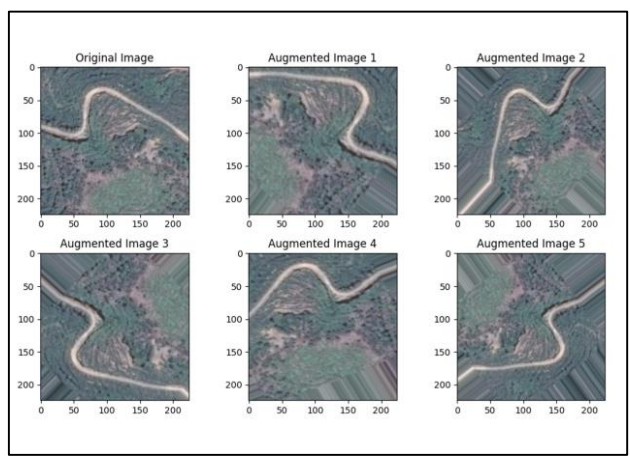

**Fig. 4:** Samples of image augmentation

*Convolutional Neural Network (CNN)*

A Convolutional Neural Network (CNN) is an artificial neural network that is distinguished by its deep feed-forward architecture (Palakodati *et al.*, 2020). Convolutional Neural Networks (CNNs) are specifically designed to effectively process and analyze visual input, namely images. A Convolutional Neural Network (CNN) typically comprises convolutional layers that use sets of adaptive filters to analyze the input image. This enables the network to identify and extract significant features and patterns. In a CNN, there are typically two main sections: The feature extraction section and the Fully Connected (FC) layer section, as illustrated in Fig. (5). The feature extraction section focuses on extracting relevant characteristics derived from the supplied data. In contrast, the FC layers are responsible for the final classification. Typically, the input of a CNN generally is a 3-dimensional tensor representing the height, width, and depth of the input data. The depth dimension refers to the number of channels in the image. For instance, in an RGB image, the depth would be three, representing the red, green, and blue channels. Local operations are carried out by the CNN's convolutional filters, which determine the dot product of the input and the filter weights. Through these processes, the network is able to extract spatial information and identify important characteristics from the input picture (Alzubaidi *et al.*, 2021).

It is common practice to use pooling layers to reduce the size of the feature maps that come after the convolutional layers. Training becomes quicker and overfitting is reduced with pooling layers because they summarise the most significant data in the feature maps, reducing the spatial dimensionality (Alzubaidi *et al.*, 2021). Finally, the fully connected layers in the CNN create high-level abstractions and perform classification tasks. These layers take the flattened feature maps from the previous layers and apply weights to produce the final classification output. We have proposed a CNN model to classify the geographical land structure. The CNN architecture takes input images with dimensions of 224×224×3, representing the height, width, and depth of the photos.

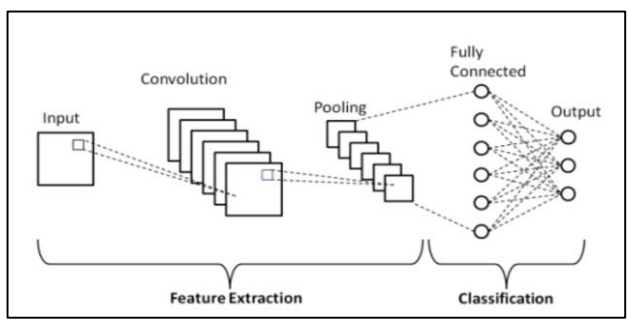

**Fig. 5:** The feature extraction section and (FC) layer section (Mohammad *et al.*, 2021)

The 1st convolutional layer of the Convolutional Neural Network (CNN) has 32 filters, each with a size of 3×3, and does not have any padding. This layer uses filters to process the incoming picture and extract crucial characteristics. Afterward, the outputs of the first convolutional layer undergo max pooling with a pool size of 2×2. The result obtained from the max pooling layer is then sent into the subsequent convolutional layer, which consists of 68 filters of size 3×3. This layer performs further feature extraction on the previous layer's outputs. The same steps are repeated for the last convolutional layer, which employs 128 filters. Each convolutional layer learns and captures different features from the input image and max pooling reduces the spatial dimensions and retains the features of the images.

Furthermore, the Glorot uniform initialization, also known as the Xavier uniform initialization, is utilized as the kernel initialization method (Glorot and Bengio, 2010). This initialization technique aims to set the initial weights of the network in a way that preserves the variance of the activations and gradients across layers during both forward and backward propagation. By maintaining a balanced initialization, the learning process becomes more stable and the convergence of the network is improved. The Glorot uniform initialization is particularly beneficial in deep learning architectures like CNNs, where the network consists of multiple layers. By initializing the weights appropriately, the network can effectively propagate information through the layers, preventing issues such as vanishing or exploding gradients. This initialization strategy enhances the overall stability and efficacy of the learning process in CNN, allowing the model to acquire and extract significant features from the input data.

In addition, the CNN design applies an activation function after every convolutional operation before sending the output tensor to the next max-pooling layer. The activation function used here is the rectified linear activation function or ReLU. A simple but powerful non-linear function, the Rectified Linear Unit (ReLU) activation function incorporates non-linearity into the network's computations. The Rectified Linear Unit (ReLU) is an activation function that efficiently overcomes the vanishing gradient problem and avoids saturation in the positive zone. It is computationally efficient. When the input is positive, the ReLU activation function returns the value of the input; when the input is negative, it returns zero, as shown in Eq. (1):

$$f(x) = max(0, x) \qquad (1)$$

After that, for classification, the CNN is linked to a fully connected layer. The second convolutional layer's feature map is first flattened, or reshaped, into a 1-dimensional vector, before it is linked to the fully connected layer. The next dense layers are able to handle the characteristics that were extracted efficiently because of this phase. The flattening procedure is followed by the use of 2 thick layers. The 1st dense layer uses the Rectified Linear Activation function (ReLU) as its activation function to link the flattened features to 64 nodes. Lastly, a softmax activation function is linked to the output of the last dense layer, which converts the numerical outputs into a probability distribution across the four classes: Terrace, meadow, farmland, and desert. The softmax activation assigns probabilities to each class, allowing the model to predict by selecting the class with the highest probability.

*Transfer Learning*

It is a method that entails using the information and insights acquired from performing one challenge and transferring them to tackle similar tasks (Sharma and Parikh, 2022). It is often used when the domains of the functions are comparable in order to have a more substantial beneficial influence (Zhuang *et al.*, 2020). Within the realm of artificial neural networks, transfer learning refers to the practice of using a pre-trained model as a starting point. The pre-trained model, with its current layers and weights, acts as a basis for the bespoke model. The pre-trained model's layers and weights are transferred to the custom model and extra custom final layers are appended to adapt the model for the job. This strategy enables the custom model to use the information and representations acquired by the pre-trained model, resulting in improved performance and increased accuracy on the target task.

Resnet-50: This residual network, often known as ResNet, was introduced by He *et al.* (2016) and is very influential in improving the training of neural networks. It represents the layers as residual functions that learn from the preceding layer. The idea of residual arises from the discrepancy between the observed and estimated values (Liang, 2020). It is a well-known example of ResNet. Included in the network's 50 layers are a max pooling layer and an average pooling layer in addition to 48 convolutional layers. The use of ResNet-50 for scene classification has been successful. On the other hand, a dense layer with four nodes that predicts the four categories of the labels using the softmax activation function has been substituted for ResNet-50's original output layer for this assignment.

Inception-v3: A deep Convolutional Neural Network (CNN) architecture that was created as a component of GoogleNet for the large-scale image visual identification challenge (Alom *et al.*, 2021). The goal of Inception-v3 was to strike a balance between computational efficiency and low parameter count, making it suitable for practical applications (Cao *et al.*, 2021). This CNN architecture, which consists of 48 layers, was trained on the ImageNet dataset, which comprises over 1 million images. In this model, the initial convolutional layer receives input data with dimensions of 299×299×3 and produces an output tensor of size 8×8×2048. The inception-v3 model has also been used to identify the scenes or landscape. Similar to the approach adopted for ResNet-50, the original output layer of Inception-v3 has been replaced with a dense layer comprising four nodes. This modification allows the model to generate predictions for the four categories of labels in the dataset. Furthermore, the input layer has been adjusted to accommodate images with dimensions of 224×224×3, aligning with the specific requirements of the dataset3.

*Phase 4: Optimization Algorithms*

An optimizer is a technique or approach used in deep learning to train a neural network to minimize the loss function by adjusting its parameters (Mehmood *et al.*, 2023).

*ADAM*

Adaptive Moment Estimation (ADAM) is an optimization technique that is often used for training deep learning models. It combines RMSProp and AdaGrad, two more optimization approaches, into one. The efficiency, quick convergence and handling of sparse gradients that ADAM is famous for are well-known (Reyad *et al.*, 2023). The main characteristics of ADAM include parameter-specific adaptive learning rates, direction-bias correction for estimation biases, $L^2$ regularization to avoid overfitting, and momentum to accelerate appropriate gradients. For these reasons, ADAM is a go-to tool for improving neural networks across the board regarding deep learning.

*SGD*

Stochastic Gradient Descent (SGD) is a crucial optimization method for training ML models, including deep NNys. The SGD algorithm iteratively changes the model parameters by using the gradients of the loss function with respect to those parameters (Tian *et al.*, 2023).

*RMSProp*

Root mean square propagation, or RMSProp, is a popular optimization method for training deep-learning neural networks. Because it adjusts the learning rate for every parameter, RMSProp overcomes some of the shortcomings of the standard Stochastic Gradient Descent (SGD) approach (Hong and Chen, 2024). Optimizing learning rates according to historical gradients aids in neural network training and is an efficient optimization strategy. It has proven helpful in several neural network designs and is extensively included in deep learning frameworks.

*Phase 5: Performance Metric*

The testing step is all about seeing how well the CNN models do. The accuracy of a Convolutional Neural Network (CNN) is one of the most common metrics used to assess its performance. Relative to the total number of photos, accuracy measures how many predictions were right. The equation for accuracy may be expressed as Eq. (2):

$$Acc(\%) = \frac{TP + TN}{TP + FP + TN + FN} \times 100\% \quad (2)$$

In this context, *TP* indicates that the model properly assigns a class to a sample, and *TN* indicates that the model correctly assigns no class to the sample, If the model gets a sample's class membership prediction wrong (*FP*), then the sample is definitely part of that class; if it gets the class membership prediction wrong (*FN*), then the sample is part of that class.

## Results and Discussion

In order to fine-tune and validate the model, we have chosen several optimizers and learning rates. These optimizers include the Adam optimizer, Stochastic Gradient Descent (SGD) optimizer, and RMSprop optimizer. The learning rates selected for training are 0.001 and 0.0001. Each combination of optimizer and learning rate is used to train the model. The training process is conducted for a fixed number of epochs, specifically 100 epochs, with a batch size of 64. These models will be trained, tested, and validated on the same sample of the dataset, in which the training set consists of 8,400 images (60%), the testing set contains 4,200 images (30%) and the validation set comprises 1,400 images (10%).

*Results*

We will show all the results obtained from our study.

*Convolution Neural Network(CNN)*

Figure (6) illustrates the training history of a CNN model utilizing the RMSProp optimizer with a learning rate of 0.0001. The plot showcases the progression of training and validation accuracy throughout the epochs. Initially, both training and validation accuracy steadily improve, demonstrating the model's capacity to acquire knowledge and apply it to new situations based on the training data. However, as the training progresses, the rate of improvement in training accuracy begins to slow down, reaching a plateau around the 50-epoch mark. Regarding training loss and validation loss, both metrics exhibit a consistent decrease during the training process.

This suggests that the model is effectively minimizing the discrepancy between predicted and actual labels. However, it's worth noting that the validation loss experiences sporadic spikes, particularly in the range of 0.2-0.4, starting from epoch 30 until the end of training. It is essential to mention that the CNN model did not train for the entire 100 epochs as specified. The early stopping mechanism was implemented during training, monitoring the model's progress based on the validation loss. After a certain amount of epochs, if the loss of validation has not improved, the training phase is ended too soon. In order to avoid overfitting and maximize training time, this method stops training the model when its validation results don't improve.

Table (1) presents the training results of the CNN model using different combinations of optimizer and learning rates. The findings indicate that CNN performs better with a lower learning rate of 0.0001, regardless of the optimizer used. Among the models trained with various parameters, the CNN model trained with RMSProp optimizer stands out with a test accuracy of 94.8% and a test loss of 0.1727. On the other hand, the CNN model trained with SGD optimizer did not perform well in this case, exhibiting the lowest results with a test accuracy of 61.3% and a test loss of 1.0444 when the learning rate was set to 0.001. These observations suggest that a lower learning rate facilitates better performance for the CNN model in this scenario and the choice of optimizer also plays a significant role in achieving optimal results.

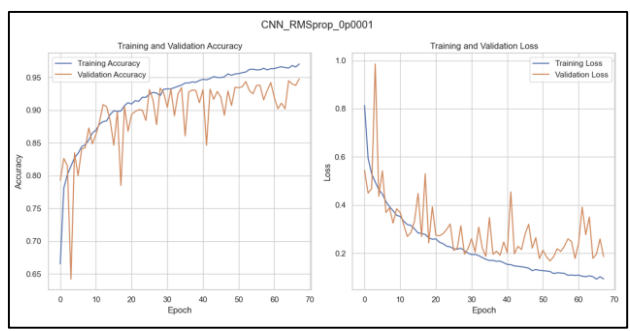

**Fig. 6:** Training history of a CNN model

**Table 1:** Performance CNN with optimizers

| Model | Optimizer | Learning rate | Accuracy % | Loss |
|---|---|---|---|---|
| CNN | ADAM | 0.001 | 92.1 | 0.2331 |
|  |  | 0.0001 | 93.7 | 0.2079 |
|  | SGD | 0.001 | 61.3 | 1.0444 |
|  |  | 0.0001 | 74.9 | 0.6328 |
|  | RMSProp | 0.001 | 92.1 | 0.2530 |
|  |  | 0.0001 | 94.8 | 0.1727 |

*ResNet-50 Model*

Table (2) shows the performance of ResNet-50 trained with different parameters. In contrast to the CNN model, the impact of a low learning rate on the ResNet-50 model's performance is not as significant. However, it is worth noting that using a low learning rate in combination with the RMSProp optimizer may result in poorer performance. Among the ResNet-50 models trained with various parameters, the model trained with the Adam optimizer, with a learning rate of 0.0001, performs better than other models trained. It achieved a test accuracy of 77.3% and a test loss of 0.6219. On the other hand, the ResNet-50 model trained with the SGD optimizer exhibited the lowest performance, with a test accuracy of 45.3% and a test loss of 1.1683. This finding shows that the impact of the learning rate is lesser compared to CNN and the selection of an appropriate optimizer remains crucial in achieving optimal results.

*Inception-v3 Model*

Table (3) shows the training performance of Inception-v3 under different parameter configurations. Interestingly, the performance of the three optimizers is quite similar in this case. The Inception-v3 model trained with the RMSProp optimizer and a learning rate of 0.0001 achieved the best results, with a test accuracy of 93.8% and a test loss of 0.1849. Close behind, the model trained with the Adam optimizer, with a learning rate of 0.0001, achieved a test accuracy of 93.7% and a test loss of 0.1867. Although slightly lower, it is a competitive performance. Furthermore, the performance gap between the SGD optimizer and the other optimizers has narrowed. The SGD model trained with a learning rate of 0.0001 achieved an 85.7% test accuracy and a test loss of 0.4225. While still not performing as well as the other optimizers, this demonstrates a relatively improved performance for the SGD optimizer with Inception-v3.

*Discussion*

To summarise the previous results, Fig. (7) presents a bar chart comparing the performance of the best model from each of the different algorithms. The chart clearly demonstrates that the CNN model utilizing the RMSProp optimizer and a learning rate of 0.0001 performs better than the ResNet-50 and Inception-v3 models. It achieved the lowest test loss (0.17) and the highest test accuracy (95%). Based on these findings, the suggested CNN model is deemed better appropriate for identifying scenes in the current experiment.

Following the identification of the optimal model a CNN trained using the RMSProp, the optimizer, and a rate of learning of 0.0001 further analysis was carried out. The top model's categorization report is shown in Fig. (8). The results demonstrate that the trained model did an excellent job, with high F1 scores, recall, and accuracy in every class. All of the classes have very high accuracy ratings, between 0.92-0.98. What this means is that the model is generally accurate when it predicts that a picture belongs to a particular class. All of the courses had excellent recall ratings, too, between 0.92-0.98. This proves that the model is accurate in identifying most occurrences of each class. All classes show good performance for the model with F1-scores that range from 0.93-0.98, which takes accuracy and recall into account. This points to the model's dependable overall performance, which is likely the result of a balanced trade-off between recall and accuracy.

**Table 2:** Performance ResNet-50 WIT H optimizers

| Model | Optimizer | Learning Rate | Accuracy % | Loss |
|---|---|---|---|---|
| ResNet-50 | ADAM | 0.0010 | 76.5 | 0.6137 |
|  |  | 0.0001 | 77.3 | 0.6219 |
|  | SGD | 0.0010 | 45.3 | 1.1683 |
|  |  | 0.0001 | 50.3 | 1.3311 |
|  | RMSProp | 0.0010 | 75.0 | 0.6712 |
|  |  | 0.0001 | 65.2 | 0.8243 |

**Table 3:** Performance Inception-v3 with Optimizers

| Optimizer | Learning Rate | Accuracy % | Loss |
|---|---|---|---|
| ADAM | 0.001 | 93.5 | 0.1958 |
|  | 0.0001 | 93.7 | 0.1867 |
| SGD | 0.001 | 91.9 | 0.2350 |
|  | 0.0001 | 85.7 | 0.4225 |
| RMSProp | 0.001 | 93.2 | 0.2048 |
|  | 0.0001 | 93.8 | 0.1849 |

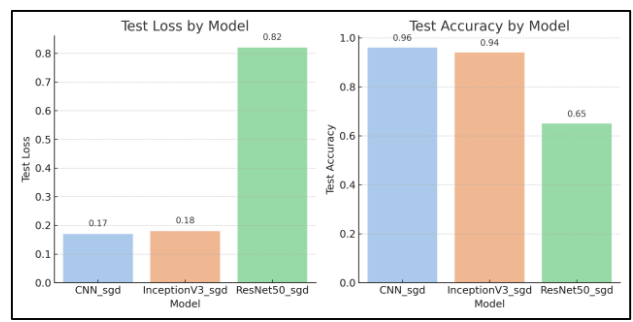

**Fig. 7:** Comparing the performance of the best model

```
              precision    recall  f1-score   support

     desert       0.98      0.98      0.98      1050
   farmland       0.92      0.94      0.93      1050
     meadow       0.95      0.92      0.93      1050
    terrace       0.95      0.95      0.95      1050

   accuracy                           0.95      4200
  macro avg       0.95      0.95      0.95      4200
weighted avg      0.95      0.95      0.95      4200
```

**Fig. 8:** Result of the best model

Figure (9) displays the confusion matrix, providing a visual representation of the model's predictions. The confusion matrix reveals the model's performance in correctly predicting the classes or labels of the images. Upon examining the matrix, it is evident that the model performs exceptionally well, with a high number of images correctly predicted across all classes. The number of images correctly predicted ranges from 962-1029 out of 1050 images for each class. However, it is noteworthy that there is a slightly higher number of wrong predictions between the" meadow" and" farmland" classes compared to the other courses. This observation aligns with our understanding that these two classes can exhibit visual similarities in the real world, making them more challenging to differentiate.

Figure (10) presents the predictions made by the CNN model on a selection of samples from the testing set. Ten samples were selected for this particular analysis to assess the performance of the model. Remarkably, the model excels by correctly predicting all the classes or labels assigned to the images. This outcome demonstrates the high accuracy and effectiveness of the proposed CNN model in accurately detecting and classifying scenes or geographical land.

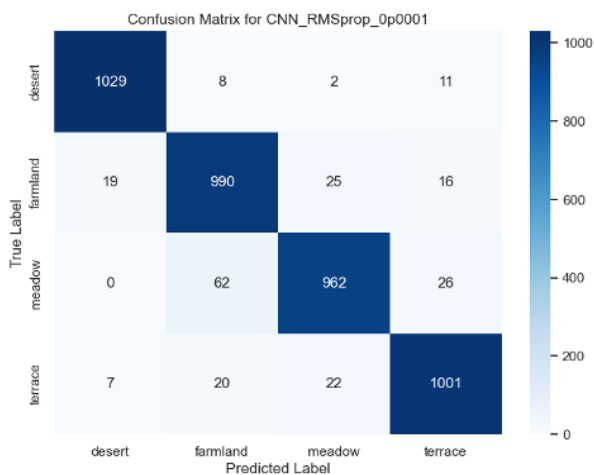

**Fig. 9:** Confusion matrix

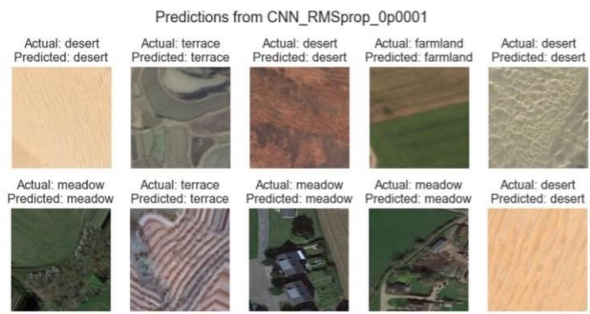

**Fig. 10:** Predictions by the CNN with RMSprop

## Conclusion and Future Work

This study aimed to develop a deep learning model for the classification of geographical land structures with three different architectures, namely CNN, ResNet-50, and Inception-v3, to identify the most suitable model for this task. The experimental results revealed that the CNN model trained with RMSProp optimizer and a learning rate of 0.0001 reached the highest level of performance with the lowest test loss (0.17) and the highest test accuracy (95%) among all the tested models. Furthermore, an in-depth analysis of the CNN model showcased its excellent precision, recall, and F1 scores across all classes, indicating its robustness and reliability in accurately identifying different land structures. Consequently, it may be said that the suggested CNN model outperformed ResNet-50 and Inception-v3 for the specific task of scene detection. This study demonstrates the efficacy of deep learning models in scene understanding. It showcases the potential for their application in various real-world scenarios related to land structure identification and classification.

In the future, the trained model may be used for real-world scenarios, including the identification and classification of land formations. This might include automating processes like environmental monitoring, urban planning, and land cover mapping via the use of remote sensing platforms or Geographic Information Systems (GIS).

## Acknowledgment

Thank you to everyone on the Editorial board who helped with the study's evaluation and editing. I would like to express my gratitude to the publisher for their assistance in publishing this essay.

## Funding Information

This study was funded by the Islamic University in Najaf, Iraq.

## Author's Contributions

**Mustafa Majeed Abd Zaid:** Conducted a comprehensive assessment of the existing literature, made significant contributions to the construction of the deep learning models, and carried out the experiments. Contributed to the gathering and preparation of data.
**Ahmed Abed Mohammed:** Initiated and formulated the study, devised the research technique, and oversaw the whole project. Tasked with the responsibility of doing data analysis and interpreting the obtained findings.
**Putra Sumari:** Provided critical feedback on the study design and methodology. Assisted in the interpretation of results and contributed to the revision of the manuscript.

## Ethics

There has been no prior publication of this material; it is completely original. There are no ethical concerns, according to the corresponding author and all authors have seen and authorized the work.